# EEG Representation Using Multi-instance Framework on The Manifold of Symmetric Positive Definite Matrices for EEG-based Computer Aided Diagnosis

Khadijeh Sadatnejad, Saeed S. Ghidary, Reza Rostami, and Reza Kazemi

*Abstract*—The generalization and robustness of an electroencephalogram (EEG)-based computer aided diagnostic system are crucial requirements in actual clinical practice. To reach these goals, we propose a new EEG representation that provides a more realistic view of brain functionality by applying multi-instance (MI) framework to consider the non-stationarity of the EEG signal. The non-stationary characteristic of EEG is considered by describing the signal as a bag of relevant and irrelevant concepts. The concepts are provided by a robust representation of homogenous segments of EEG signal using spatial covariance matrices. Due to the nonlinear geometry of the space of covariance matrices, we determine the boundaries of the homogeneous segments based on adaptive segmentation of the signal in a Riemannian framework. Each subject is described as a bag of covariance matrices of homogenous segments and the bag-level discriminative information is used for classification. To evaluate the performance of the proposed approach, we examine it in attention deficit hyperactivity/bipolar mood disorder detection and depression/normal diagnosis applications. Experimental results confirm the superiority of the proposed approach, which is gained due to the robustness of covariance descriptor, the effectiveness of Riemannian geometry, and the benefits of considering the inherent non-stationary nature of the brain.

*Index Terms*— computer aided diagnosis, covariance matrix, multi-instance learning framework, non-stationary EEG signal, Riemannian manifold.

## I. Introduction

In this paper, we propose a new representation of EEG signal to be used in EEG-based computer aided diagnosis (CAD) systems to diagnose psychiatric disorders. Routinely, the mental disorders are diagnosed by qualitative diagnostic criteria, such as diagnostic and statistical manual of mental disorders [1] used in clinical interview, which can be impressed by physician knowledge and attitude, with a relatively low agreement in different clinicians' decision. Diagnosis using these behavioral observations has been often unreliable and is highly dependent on the clinicians' intuition. Early and accurate diagnosis of different mental disorders is important for proper treatment approach selection and prevention of mortality and morbidity.

EEG-based CAD as a quantitative method for diagnosis, which relies on analysis of electrical activity of the brain, can help clinicians to increase the confidence of diagnosis. Recently, numerous studies have investigated this problem from different points of views. However, the generalization and robustness of those systems are still crucial bottlenecks. These two features can be considered in each component of a diagnostic system, including preprocessing, feature extraction, and classification components.

In a general categorization, EEG-based CAD systems represent data using linear and non-linear approaches. Although some researches that use linear approach have shown promising results in diagnosing mental disorders [2, 3, 4, 5]; however complex, non-linear, and non-stationary nature of EEG signals inevitably leads to insufficiency of linear methods for representing EEG signals [2, 52]. Different nonlinear methods such as power spectrum, wavelet entropy and energy, phase entropies, approximate entropy, fractal dimension, and correlation dimension have been used for representing EEG signals in different diagnostic applications, including seizure detection, diagnosis of epilepsy [6, 7, 8, 9, 10], Alzheimer [11], schizophrenia [12], and depression [13, 14, 15]. The promising results achieved by these methods confirm the suitability of non-linear approaches for representing EEG in EEG-based CAD systems.

Recently a new approach for EEG signal analysis based on the Riemannian geometry has been developed [16], which uses the spatial covariance matrix of EEG recording as a non-linear representation of EEG signals. This method has gained promising results in EEG signal analysis applications, especially in brain computer interfacing (BCI). Other researches also have reported reliability, robustness, and good performance for considering the Riemannian geometry of symmetric positive definite matrices [17]. These findings advocate us to use the spatial covariance matrix of EEG segments for representing the recordings in an EEG-based CAD system.

The analysis of covariance matrices or in general terms, statistical data processing approaches for EEG signal analysis is based on the assumption of stationarity of the signal, while the brain activity is essentially non-stationary [18]. A common approach to resolve this conflict is dividing the EEG signal into short time segments to satisfy piecewise stationary condition (i.e. fixed-size segmentation approach) [19]. Analysis of these segments may lead to some practical insight, however, it is statistically inefficient and provides an incomplete representation of the EEG signal [18]. This shortcoming is resulted from the trade-off between the length of time segments and the stationarity assumption. For example, in the case of covariance matrices being used as descriptors, analysis in very short time segments would lead to indefinite sample covariance matrices, while in longer time segments it would lead to the occurrence of heterogeneities within segments with higher probability. Dividing EEG signals into the homogeneous pieces using adaptive segmentation methods [19], fairly resolve this problem by determining the length of homogeneous segments adaptively.

In contrary to the usual approaches that consider non-stationarity to be the result of external stimuli on brain functioning, the non-statinarity can be considered as the result of switching of the metastable state of the neural assemblies [18] or pathological changes. Ignoring these significant sources of non-stationarity can result in missing some valuable information of the brain functionality or loss of confidence in EEG-based diagnosis systems.

It is noteworthy that all derived homogenous segments are not necessarily relevant to the subject's class label, (e.g. regions dominated by noise, different artifacts, biological functioning). When the spatial covariance matrices are used for describing the segments of EEG signal and a Riemannian metric is used to compute the distances [17], we can assume artifacts, noise, and any kind of brain activity that is not related to the mental disorder to have representations that are adequately different than the representation of patterns appeared by the mental disorder in EEG signal. Therefore, it could be expected to have a feature space with multi-modal distribution of data points to represent a subject. If we assign the same label to all pieces extracted from the signal recorded for a person and then embed all of the derived homogenous segments in a single instance learning framework, it would result in a complex and non-linear separable distribution of data points (i.e., with overlapping between two classes in some parts).

Considering the above mentioned facts, our aim in this study is to introduce a new representation which has three simultaneous objectives; 1) including the non-stationarity of the EEG signal for describing it to reflect the importance of the sources which lead to non-stationarity of the signal in mental disorder diagnosis, 2) overcoming the shortcoming of fixed-size segmentation using adaptive segmentation and 3) overcoming the shortcomings of single-instance learning process using multi-instance learning framework.

In this new representation, we describe the extracted segments using a robust representation such as covariance matrix, then the resulting relevant or irrelevant patterns are embedded in a multi-instance (MI) framework. This representation is conceptually compatible with the heterogeneity of EEG signal during recording, which is produced by different sources of non-stationarity. Therefore, a subject is described using a bag of concepts (representation of homogenous intervals) in the MI framework [20]. The concepts are derived by applying an adaptive segmentation approach to avoid the shortcomings of using fixed-size segmentation of EEG signals. The robust representation of homogenous segments leads to the robustness of the diagnostic system. This representation is based on applying Riemannian geometry for determining the boundaries of homogeneous segments and describing the dataset as a similarity matrix using MI kernel.

The remainder of the paper is organized as follows: In section II we describe the mathematical preliminaries which are required for better understanding of this paper. In section III our proposed approach for representing and analyzing EEG signal is described. Experimental setups and the results are described in section IV. Finally, the results are concluded in section V.

## II. BACKGROUND

In this section we first review some basic concepts in Riemannian geometry, then describe preliminaries of multi-instance learning framework, which are necessary for reading the paper.

### A. Riemannian geometry and the manifold of SPD matrices

A topological manifold is a connected Hausdorff space that for every point of the manifold, there is a neighborhood $U$, which is homeomorphic to an open subset $V$ of $\mathbb{R}^d$ but globally has a more complex structure. The homeomorphism between sets U and V ($\varphi$: U→V) is called a (coordinate) chart. A family of charts that provides an open covering of the manifold is called an atlas, $\{(U_\alpha, \varphi_\alpha)\}$. The pairs $(U_\alpha, \varphi_\alpha)$ are the local charts or coordinate systems. A differentiable manifold is a manifold with an atlas, such that all transitions between coordinate are smooth.

A Riemannian manifold $(M, g)$ is a differentiable manifold $M$, which is endowed with a smooth inner product (Riemannian metric $g(u,v)$) on each tangent space $T_X M$. The inner product (Riemannian metric) in Riemannian manifolds is a metric that allows measuring similarity or dissimilarity of two points on the manifold [21, 22, 23].

A curve $\gamma: I \subset R \to M$ is a geodesic if the rate of change of $\dot{\gamma}$ has no component along the manifold for all $t \in I$ or $\ddot{\gamma}$ is 0 [21]. Given a vector $v$ in the tangent space $T_X M$, there is a geodesic $\gamma(t)$ which is characterized by its length, where geodesic issued from $\gamma(0) = X$, and $\dot{\gamma} = v/\|v\|$. Two points on the manifold may have multiple geodesic between them, but the ones which have minimum length is called minimizing

geodesic [21].

In this paper, we describe the inputs of our diagnostic system using the covariance matrices. The space of covariance matrices (symmetric positive definite matrices) does not satisfy the scalar multiplication axiom of a vector space (i.e. the multiplication of an SPD matrix with a negative scalar value is not an SPD matrix). Since the space of $d \times d$ dimensional SPD matrices, Symd+, forms a convex cone in $\mathbb{R}^{d^2}$ Euclidean space, using a Riemannian metric to analyze the geometry of the space of $Sym_d^+$ is more compatible with its non-linear structure in comparison with investigating it in $\mathbb{R}^{d^2}$ Euclidean space. A number of different metrics have been proposed for $Sym^d_+$ to capture its non-linear structure [17, 24, 25]. For example, log-Euclidean [24] and affine-invariant Riemannian metric [25] which induce log-Euclidean and affine-invariant geodesic distances are two popular metrics used over the manifold of SPD matrices. A geodesic that connects two SPD points in log-Euclidean framework is defined as:

$$\gamma(t) = exp\big((1-t)\log(C_1) + t\log(C_2)\big) \quad (1)$$

where $t \in [0,1]$ and $C_1, C_2 \in Sym_d^+$. Log-Euclidean geodesic distance between $C_1$ and $C_2$ (i.e. the minimizing geodesic derived from log-Euclidean metric) can be expressed as:

$$d_{LE}(C_1, C_2) = \|log(C_1) - log(C_2)\|_F \quad (2)$$

where $d_{LE}(C_1, C_2)$ is the log-Euclidean distance and $\|.\|_F$ denotes the Frobenius matrix norm. The affine-invariant metric is the other effective metric which induces geodesic distance over the manifold of SPD matrices:

$$d_{AI}(C_1, C_2) = \left\|log\big(C_1^{-1/2} C_2 C_1^{-1/2}\big)\right\|_F \quad (3)$$

where $d_{AI}$ denotes the affine-invariant geodesic distance between two $C_1, C_2 \in Sym_d^+$ points [24].

*B. Multi-instance learning framework*

Multiple-instance (MI) learning [20] is a variety of inductive machine learning methods (commonly supervised learning), which instead of learning over a set of individually labeled instances, the learner receives sets of labeled bags. For example, in the multiple-instance binary classification, let $\chi$ be the instance space (or the space of feature vectors) and $\Omega = \{+,-\}$ be the binary class attributes. The aim of learning in MI framework is finding a $\nu_{MI}: \mathbb{N}^\chi \to \Omega$ function using training samples, where $\mathbb{N}^\chi$ refers to all multi-sets of $\chi$. In this framework, each bag is composed of one or more occurrences of different instances.

Standard MI assumption supposes that each instance belonging to a bag has a hidden label which is either positive or negative $c \in \Omega = \{+,-\}$. A positive label can be assigned to a bag if and only if it contains at least one positive instance. In other words, let $X = \{x_1, x_2, ..., x_n\} \in \mathbb{N}^\chi$ be a bag or multi-set with $n$ instances. The label of an instance is defined by a function $h: \chi \to \Omega$. Considering the positive and negative labels in concept-level as logical "true" and "false" constants, we can state the standard MI assumption as:

$$\nu_{MI} \Leftrightarrow (h(x_1) \vee h(x_2) \vee ... \vee h(x_n)) \quad (4)$$

where $\nu_{MI}$ is the MI concept function. Several single-instance learning methods such as support vector machine (SVM) [26], neural networks [27], decision trees [28, 29], and ensemble-based learning [30] have been adapted to the multi-instance framework under the standard MI assumption.

In MI learning literature, different relaxations of this hard assumption are proposed to satisfy the requirements of some other problems, which are not exactly compatible with the standard MI assumption. For example, metadata-based approach is a simple MI learning approach which replaces each bag with a metadata feature vector derived somehow from the instances of that bag. Therefore, the MI learning problem is converted to a single instance learning problem. This transformation can be done either implicitly or explicitly. After this transformation, in the learning process, a single-instance learning algorithm can be applied to the resulting transformed feature space. This approach implicitly is based on an assumption, called metadata assumption, which states that the labels of the samples are directly related to their representation in transformed space (metadata feature space) [47]. For example, [31] proposed MI kernels that can apply any kernel-based method such as standard SVM algorithm to MI data. The kernel is defined as:

$$k_{MI}(X, Y) = \sum_{x \in X, y \in Y} k_I^p(x, y) \quad (5)$$

where $k_I^p$ is the $p$th power of an instance-level kernel $k_I$, $x$ and $y$ denote instances while $X$ and $Y$ denote subjects. For a positive definite (*pd*) kernel $k_I$, the *p*th power of it generates a *pd* kernel and it is proved that for $p$ values that are sufficiently large, if the instance-level kernel is separable its corresponding MI kernel is separable too.

### III. PROPOSED METHOD

We propose a new representation of non-stationary EEG signal, which intends to be robust and leads to an appropriate generalization of the diagnosis system (Fig. 1). In this section, we first introduce our representation and then describe an EEG segmentation method.

*A. MI representation of EEG signal*

The psychiatric disorders are commonly used as informative sources of non-stationarity of the EEG signal in CAD applications. They might be diagnosed through the detection of specific patterns in patient's brain electrical activity. The other major sources of non-stationarity are ambient noise, artifacts, and biological sources, which inevitably affect the signal during recording. To represent the varieties produced by different sources of non-stationarity, we interpret the recorded trial of subject $X$, as a bag of different concepts ($X = \{x_1, ..., x_i, ..., x_n\}, x_i \in \chi$). Here concepts are the covariance matrices of homogeneous segments (in statistical sense), resulted from adaptive segmentation of EEG signals (section III. B). The covariance matrices are approximated using sample covariance matrices as:

$$C(t,T) = \frac{1}{T-1} E_{(t,T)} * E_{(t,T)}^T \quad (6)$$

where $E_{(t,T)}$ denotes the EEG recording started at second $t$ of a segment with the total length of $T$ seconds, and $C(t,T)$ denotes the resulting covariance matrix.

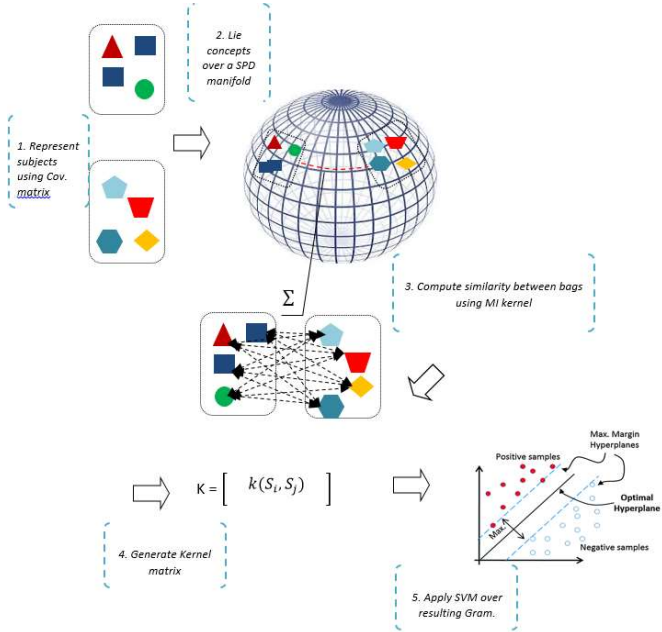

Fig. 1. The steps of proposed CAD system.

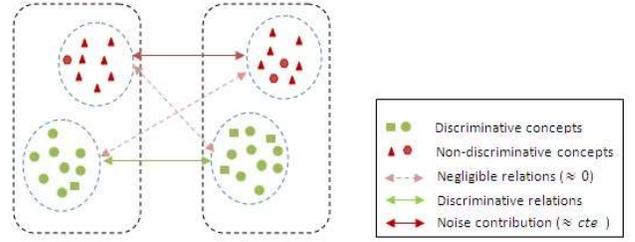

Fig. 2. Different types of relations in computing MI kernel between two bags

Different sources of non-stationarity of EEG signal causes some concepts to be relevant and some others to be irrelevant to the class label. We call the relevant ones positive (+) concepts and the irrelevant ones negative (−) concepts. In an EEG-based diagnosis system, obviously a subject can be labeled with a specific disorder if it contains the patterns which are corresponding to that disorder in its EEG signal (positive instances corresponding to that class). This representation and interpretation of recorded non-stationary trials are compatible with the multi-instance learning assumption. Therefore, by embedding the retrieved instances in MI framework, we can consider the discriminative information on bag-level using MI learning algorithms.

In some mental disorders the patterns of the disorders are easily recognized by experts via visual inspection of the patient's brain electrical activity. However, in some others more accurate analysis of the EEGs is required to detect the abnormalities of the signal affected by brain disorders. In these cases, explicit concept-level labeling of the signals is impossible and it is required to adapt machine learning techniques for mining the underlying patterns. As the MI kernel approach is able to compute similarities without explicit inducing the concepts, we use it for EEG signal representation.

In MI kernel approach the similarity between subjects is computed by a generalization of the inner product between every two concepts obtained from distinct subjects as:

$$k(X,Y) = \frac{\sum_{x_i \in X, y_j \in Y} k_\chi(x_i, y_j)}{N*M} \quad (7)$$

where $k(X,Y)$ denotes the similarity between $X$ and $Y$ bags that represents two subjects, $x_i$ and $y_j$ denote the concepts that belong to the $X$ and $Y$ bags, respectively, and $N$ and $M$ denotes the number of concepts in the $X$ and $Y$ bags. $k_\chi(x_i, y_j)$ is an instance-level kernel defined in instance space $\chi$. According to the (7), computing MI kernel $k(X,Y)$ leads to computing $N \times M$ concept-level kernels. By considering that each subject is represented by positive and negative concepts, as illustrated in Fig. 2, these $N \times M$ calculations can be categorized into three groups, according to the types of concepts involved in computation: Computing $k_\chi(x_i, y_j)$ between two positive instances, two negative instances, and a positive and a negative instances (abstracted as discriminative relation, noise contribution, and negligible relations in Fig. 1). From the machine learning point of view the discrimination between two classes is based on the classification between their positive instances. Therefore, we consider the relations between instances of two subjects as discriminative relations. All pairs dominated by other sources of non-stationarity of the signal, which are not informative for discrimination, are considered as noise contribution. In other words, $k(X,Y)$ can be considered as the sum of three types of similarities:

$$k(X,Y) = k_{+,+}(X,Y) + k_{+,-}(X,Y) + k_{-,-}(X,Y) \quad (8)$$

$$k_{+,+}(X,Y) = \sum_{x_i \in X, y_j \in Y} k_\chi(x_i, y_j) \text{ where}$$
$$(h(x_i) = '+' \ \& \ h(y_j) = '+')$$

$$k_{+,-}(X,Y) = \sum_{x_i \in X, y_j \in Y} k_\chi(x_i, y_j) \text{ where}$$
$$(h(x_i) = '-' \ \& h(y_j) = '+') \ | \ (h(x_i) = '+' \ \& \ h(y_j) = '-')$$

$$k_{-,-}(X,Y) = \sum_{x_i \in X, y_j \in Y} k_\chi(x_i, y_j) \text{ where}$$
$$(h(x_i) = '-' \ \& \ h(y_j) = '-')$$

By considering the negative (non-discriminative) concepts (i.e. noise, artifact,..) as any kind of concepts which are different enough to be compared to the positive (discriminative) concepts [31], $k_{+,-}(X,Y)$ would have a negligible contribution in computing the similarity between every two subjects (i.e. $k_{+,-}(X,Y) \cong 0$).

Since all signals are recorded in the same resting conditions, it is not far-fetched that the similarity between different non-informative segments of the EEG signal (i.e. noise contribution) can be considered as a relatively constant offset (i.e. $k_{-,-}(X,Y) \cong cte$).

The $k_{+,+}(X,Y)$ term denotes the similarity between positive instances of two subjects and represents the discriminative information. It is expected that the subjects with a similar label (similar disorder) will show higher $k_{+,+}(X,Y)$ (because the concepts evolved from the same psychiatric disorders would be similar) while patients with dissimilar disorders will show smaller $k_{+,+}(X,Y)$ (concepts evolved from different psychiatric disorders are dissimilar). This results in higher/lower $k(X,Y)$ value for similar/dissimilar disorders (i.e. $k(X,Y) \gg k(X,Z)$ where $v_{MI}(X) == v_{MI}(Y)$ and $v_{MI}(X) \neq v_{MI}(Z)$).

Since analyzing the covariance matrices as 2nd-order tensors in Riemannian framework leads to superior results in comparison with analysis of its vectorized equivalent in Euclidean space [32], we choose $k_\chi(x, x')$ to be compatible with the Riemannian geometry of the manifold of SPD matrices. Depending on the classifiers that are used for diagnosis, different types of kernel can be selected from different points of views. We have studied several kernels which are compatible with the manifold of SPD matrices:

*1) Log-Euclidean & affine-invariant Gaussian kernel*

The Log-Euclidean kernel is defined as:

$$K_{LE}: Sym_d^+ \times Sym_d^+ \to \mathbb{R}$$
$$k_{LE,\chi}(x,y) = exp(-\frac{d_{LE}^2(x,y)}{\sigma}) \qquad (9)$$

where $k_{LE,\chi}(x, y)$ denotes the similarity between $x$ and $y$ concepts based on the Log-Euclidean distance, $d_{LE}$ and $\sigma$ is a positive value known as bandwidth parameter. Since Log-Euclidean Gaussian kernel satisfies mercer condition for all $\sigma > 0$ [32] (i.e. it is a positive semi definite kernel), the resulting $k(X,Y)$ which would be the sum of multiple positive semi definite (*psd*) kernels is a *psd* kernel and can be used in any kernel-based learning method.

Using affine-invariant Riemannian distance leads to affine-invariant Gaussian kernel:

$$K_{AI}: Sym_d^+ \times Sym_d^+ \to \mathbb{R} \qquad (10)$$
$$k_{AI,\chi}(x,x') = exp(-\frac{d_{AI}^2(x,x')}{\sigma})$$

Although this kernel generally is not a *pd* kernel, an empirical affine-invariant Gaussian kernel might be a *pd* kernel depending on the distribution of data points and its bandwidth parameter.

Both of these kernels are able to provide discriminative projection of data points into the feature spaces, therefore they are appropriate choices for SVM classifier. SVM is based on minimizing a regularized combination of empirical and structural risk. This objective function, which can be interpreted as maximizing the margin between two classes in the feature space, leads to the robustness of the SVM classifier in EEG analysis [51].

*2) Isometric kernel*

The isometric kernel which is resulted from using the double centering method [33] over the matrix of geodesic distances is an informative kernel, especially if it is used in methods which rely on the local topology of the data points.

$$S = -(1/2)JD^2J \qquad (11)$$
$$J = I_{N \times N} - (1/N)1_N \times 1_N^T$$

where $S$ is the similarity matrix, $D$ denotes the matrix of geodesic distances, $I_{N \times N}$ is an $N \times N$ identity matrix, and $1_N$ is a column vector that it's all elements are 1. The similarity matrix resulted from Isometric mapping would not satisfy mercer conditions on manifolds with non-zero intrinsic curvature [50].

*B. Identifying the homogenous segments*

To determine the boundaries of homogenous segments, we apply an adaptive segmentation approach in which the boundaries of the segments are recognized as points where the resulting segments satisfy stationarity condition (being homogeneous in the statistical sense). At first, the EEG signals are divided into elementary segments using fixed-size segmentation approach. We describe the segments using spatial covariance of the pieces (6). To detect the boundaries of homogeneous segments we compare spatial covariance matrices of the successive elementary segments in terms of geodesic distance. By applying a thresholding method, we can detect the local peaks that are identified as cut points or the boundaries of homogeneous segments of the recorded EEG. New segments will be generated from merging the successive elementary segments between two cut points. This process can be summarized as:

$$d_G\big(C(t,T), C(t+T,T)\big) > th =$$
$$\begin{cases} t+T \text{ is an interrior point of a segment} & \text{if No} \\ t+T \text{ is a cut point} & \text{if yes} \end{cases} \qquad (12)$$

where the $th$ is an empirical threshold level, and $d_G$ denotes the geodesic distance.

IV. EXPERIMENTS

In this section, we evaluate the proposed representation in two different psychiatric disorder diagnosis applications: Attention Deficit Hyperactivity Disorder (ADHD) / Bipolar Mood Disorder (BMD) and depression/ normal group.

We first describe the datasets and the recording conditions of EEG signals for these two datasets, then we describe different experimental setups and results, and finally discuss the results.

*A. Data sets*

The specification and recording conditions of the datasets are described in details as follows:

*1) ADHD/BMD [36]*

ADHD and BMD are psychiatric disorders with similar clinical symptoms. The overlap between clinical symptoms of these two disorders leads to unreliability of qualitative diagnosis approach. This dataset consists of the EEG signal of 43 children and adolescents, with 21 subjects with ADHD (age range: 10–20, age mean ± std: 14.36 ± 2.9) and 22 subjects with BMD (age range: 13–22, age mean ± std: 16.50 ± 2.50). Data acquisition was performed in the Biophysics Laboratory of Shiraz University of Medical Sciences. The children, adolescents and at least one of their parents were interviewed using DSM_IV criteria for diagnosis. It should be noted that most of the subjects were followed up by the psychiatrist for at least 6 months and the psychiatrist was assured about their type of illness. For each patient, the EEG signals were recorded in two eyes-open and eyes-closed resting conditions for 3 minutes, in order to analyze the natural behavior of their EEGs. The EEG signals were recorded using the Neuroscan-LT setup (equipped with a 16bit A/D card). The signals were recorded using 22 electrodes according to the 10–20 international recording system. The scalp channels were located in the following positions: G, Fz, Cz, Pz, C3, T3, C4, T4, Fp1, Fp2, F3, F4, F7, F8, P3, P4, T5, T6, O1, O2, A1 and A2. The average value of the A1 and A2 electrodes, which were attached to the earlobes, was used as the reference. The ground electrode (G) was placed on the forehead center. The impedances of the electrodes were lower than 10 kΩ through the recording and the sampling rate of the EEG was 250 Hz.

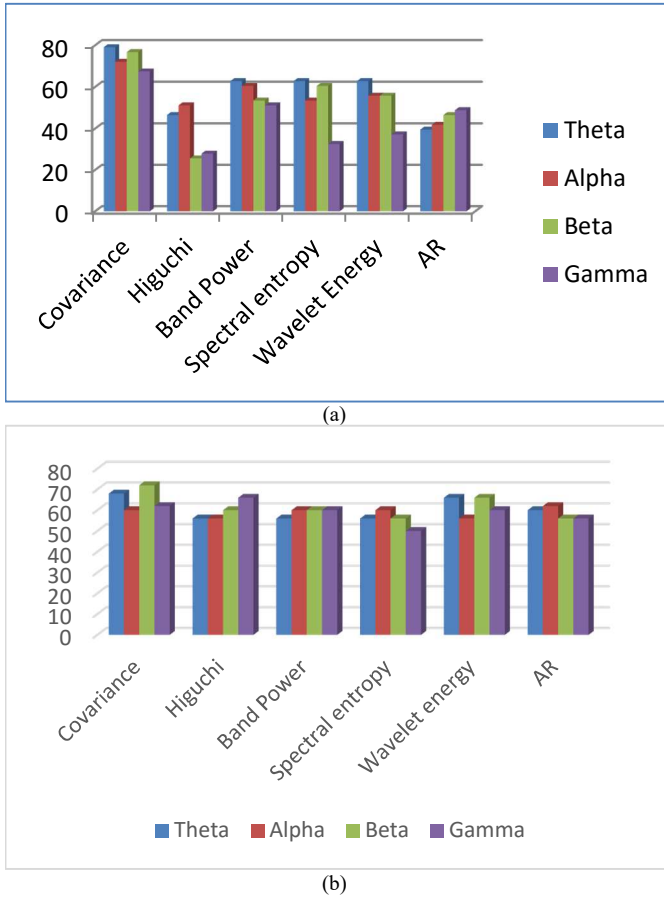

Fig. 3. Comparison of the classification accuracy of different features extracted from EEG signals of the (a) ADHD/BMD (b) depression/normal datasets in eyes-closed resting condition.

*2) Depression/Normal [37]*

Depression is a common disorder with a high prevalence of critical outcome, such as suicidal attempt. The World Health Organization (WHO) has predicted that the depression would be the second largest burden of disease by 2020 [38]. Fifty patients with bipolar depression and major depressive disorder (MDD) were recruited for this study after their disorder was confirmed according to the DSM IV-TR criteria through clinical interview. The patients were examined at the Atieh Clinical Neuroscience center during a period from February to September 2014. The inclusion criteria were the following: outpatient subjects whose age ranged from 18 to 65, diagnosed as having major depression or bipolar disorder, according to the DSM IV-TR, had a Beck depression inventory (BDI-II) score of more than 14. The exclusion criteria were: personality disorder in Axis II, seizures, epilepsy in the first degree relatives, pregnancy and head trauma [37].

For recordings, a 19 ElectroCap (ElectroCap, Inc; OH) was used, with its electrodes placed on Fp1, Fp2, F7, F3, Fz, F4, F8, T3, C3, Cz, C4, T4, T5, P3, Pz, P4, T6, O1, O2 by a 10–20 system. The A1 + A2 was used for reference. The impedance of the electrodes was under 5 kΩ. EEG was recorded in 5 minutes while the subjects were in the resting state with their eyes closed and in 5 minutes for eyes open resting condition. The recording was carried out using a Mitsar system in an acoustic room. The data were converted into numbers with 500 sample rates and with a high frequency filter of 50 Hz, a low frequency filter of 0.3.

*B. Experiments & discussion*

To evaluate the proposed approach, we designed several experiments. As the first experiment, we examined the effectiveness of covariance descriptor against a wide range of descriptors commonly used in EEG-based CAD applications, including: Higuchi [39], autoregressive [40], power spectrum [41], spectral entropy [42], and wavelet energy [43] methods. ADHD/BMD and depression/normal datasets in eyes-closed resting conditions are used in this experiment. We first filter the signal into the Theta (3-8 Hz), Alpha (8-12 Hz), Beta (12-28 Hz), and Gamma (28-50 Hz) sub-bands using the 5th–order band-pass Butterworth filter. Then the above methods are used for representing the signal in segments resulted from applying fixed-size segmentation over the recorded signals. The signals are divided into overlapping windows (50% overlapping between segments) and the length of the segments is determined empirically by applying the cross-validation and investigating a wide range of values (up to 20 seconds) as window length. We consider these features as the vectors in Euclidean space. These vectors are the result of the application of above mentioned methods on all recorded channels except A1, A2, and G channels [12]. To have a quantitative comparison between different descriptors, an SVM classifier [44] with RBF kernel is used for classification. We used the LIBSVM package [45] for implementing SVM classifier. Tuning the bandwidth parameter of RBF kernel and $C$ parameter of SVM classifier is done by applying cross-validation and examining a wide range of values (i.e. $C \epsilon \{0.1, 10, \ldots, 100000\}$, $\sigma \epsilon \{0.1, 1, 10, 20, \ldots, 1000\}$) for selecting the best value over the validation set.

The covariance matrix as a descriptor was estimated empirically by applying (6) on time segments of 19 recorded channels. The setup of this experiment was the same as the above-mentioned conditions. With the exception that the analysis was based on the Riemannian geometry of the manifold of SPD matrices and affine-invariant Gaussian kernel [32] was used for computing similarities. Fig. 3 illustrates a comparison between above mentioned descriptors. It is obvious that in most sub-bands covariance matrix as a descriptor leads to superior results.

As the second experiment, we compared the proposed MI–

TABLE I
THE COMPARISON BETWEEN THE PROPOSED DESCRIPTOR (MI-ADAPTIVE-COV) WITH SOME COVARIANCE-BASED DESCRIPTORS, IN TERMS OF CLASSIFICATION ACCURACY ON (A) ADHD/BMD (B) DEPRESSION/NORMAL DATASETS IN EYES-OPEN AND EYES-CLOSED RESTING CONDITIONS.

| ADHD/BMD (Eye-open) | | | | | |
|---|---|---|---|---|---|
| | Theta | Alpha | Beta | Gamma | 3-50 Hz |
| Batch-Cov+SVM | 79.07 | 83.72 | 83.72 | 83.72 | 79.07 |
| Mean-Cov+SVM | 76.74 | 86.05 | 86.05 | 79.07 | 79.07 |
| MI-fixed-Cov+SVM | 79.07 | 86.05 | 88.37 | 81.40 | 83.72 |
| MI-Adaptive-Cov+SVM | 88.37 | 88.37 | 90.70 | 88.37 | 83.72 |
| ADHD/BMD (Eye-closed) | | | | | |
| Batch-Cov+SVM | 79.07 | 72.09 | 76.74 | 67.44 | 74.42 |
| Mean-Cov+SVM | 72.09 | 72.09 | 72.09 | 69.77 | 62.79 |
| MI-fixed-Cov+SVM | 79.07 | 76.74 | 79.07 | 81.40 | 79.07 |
| MI-Adaptive-Cov+SVM | 83.72 | 83.72 | 88.37 | 76.74 | 88.37 |

(a)

| Depression/normal (Eye-open) | | | | | |
|---|---|---|---|---|---|
| | Theta | Alpha | Beta | Gamma | 3-50 Hz |
| Batch-Cov+SVM | 70.00 | 62.00 | 72.00 | 62.00 | 60.00 |
| Mean-Cov+SVM | 62.00 | 56.00 | 66.00 | 60.00 | 56.00 |
| MI-fixed-Cov+SVM | 72.00 | 76.00 | 76.00 | 64.00 | 82.00 |
| MI-Adaptive-Cov+SVM | 72.00 | 80.00 | 80.00 | 78.00 | 86.00 |
| Depression/normal (Eye-closed) | | | | | |
| Batch-Cov+SVM | 68.00 | 60.00 | 72.00 | 62.00 | 58.00 |
| Mean-Cov+SVM | 62.00 | 68.00 | 72.00 | 68.00 | 56.00 |
| MI-fixed-Cov+SVM | 70.00 | 70.00 | 80.00 | 82.00 | 86.00 |
| MI-Adaptive-Cov+SVM | 74.00 | 74.00 | 80.00 | 84.00 | 90.00 |

(b)

based representation (MI-Adaptive-Cov) with other covariance-based approaches, applied to the segments generated by adaptive segmentation of EEG signals. The goal of these comparisons is to evaluate the different components involved in our representation. The representations of the EEG signals for methods involved in our comparisons are as follows:

1) MI framework over the segments generated by using fixed-size segmentation (MI-fixed-Cov)

2) A batch of covariances without considering the non-stationarity of the signal. The covariances are computed in segments generated by a fixed-size segmentation approach and the segments are labeled same as the subject's label (Batch-Cov)

3) The geometric mean [46] of covariance matrices where the segments are generated by the fixed-size segmentation (Mean-Cov)

We applied SVM classifier with an affine-invariant Gaussian kernel for discrimination over these features. The $\sigma$ and $C$ parameters, as described in the first experiment, were set using cross-validation. We evaluate the methods using leave-one-out cross-validation approach. The experimental evidences confirm the superiority of the MIL-Adaptive-Cov+SVM for the both ADHD/BMD and Depression/normal datasets, in two eyes-open and eyes-closed resting conditions (Table I).

TABLE II
EXAMINING THE STATISTICAL SIGNIFICANCE OF THE PROPOSED METHOD IN COMPARISON WITH ITS COMPETITORS REPORTED IN TABLE. I.

| | Depression/Normal | | ADHD/BMD | |
|---|---|---|---|---|
| | Eyes-open | Eyes-closed | Eyes-open | Eyes-closed |
| MI-Adaptive-Cov/Batch-Cov | 0.0279 | 0.0267 | 0.0029 | 0.0029 |
| MI-Adaptive-Cov/ Mean-Cov | 0.0057 | 0.0502 | 0.0190 | 0.0129 |
| MI-Adaptive-Cov/ MI-fixed-Cov | 0.0897 | 0.0249 | 0.0705 | 0.1197 |
| MI-fixed-Cov/ Batch-Cov | 0.0915 | 0.0423 | 0.2414 | 0.0972 |

We can conclude from these experiments that the superiority of the proposed method is the result of following facts:

1) Considering the non-stationarity of the EEG signal by applying the MI framework in the proposed representation. It is confirmed by (Batch-Cov+SVM/ MI-fixed-Cov+SVM) and (Batch-Cov+SVM/ MI-Adaptive-Cov+SVM) compairings.

2) Determining the boundaries of the segments adaptively, which is confirmed by (MI-fixed-Cov+SVM / MI-Adaptive-Cov+SVM) comparison.

3) Doing all of the analysis in Riemannian framework. We did some analysis over covariance matrices using Euclidean geometry which led to the overfitting to the training samples.

4) Applying appropriate representation for basic elements (concepts) in the proposed representation, confirmed by the first experiments.

To compare the significance of the superiority of the proposed approach in comparison with other covariance-based methods, we apply paired t-test over the classification rates resulted from applying MI-Adaptive-Cov for describing subjects versus Batch-Cov, Mean-cov, and MI-fixed-Cov in different sub-bands for ADHD/BMD and depression/normal datasets. The resulting p-values for comparison between MI-Adaptive-Cov versus Batch-Cov ($< 0.05$), reported in Table II, confirm the significance of the superiority of the proposed approach against classic covariance-based representation (Batch-Cov). It also has significantly superior results in most cases in comparison with Mean-Cov. To show the importance of adaptive segmentation component, we compared non adaptive MI-fixed-Cov method against Batch-Cov. As shown in Table II the non-adaptive method has poor results compared to the proposed adaptive representation.

The resulting similarity matrix of our proposed representation has to satisfy the *psd* constraint if it is going to be used as an empirical kernel matrix for a kernel-based learning systems. Using non-*psd* kernels, when the learning system minimizes the empirical risk (such as SVM) leads to a non-convex optimization problem. In such cases, the minimization of risk function is not guaranteed [49]. In dimensionality reduction problems also, the negative eigenvalues of the similarity matrix prevent from embedding the data points in a real-valued Euclidean space. The positive definiteness of our proposed representation can be controlled by the type of concept-level kernel and its parameters.

Similar to any kernel based method, the type of kernel plays an important role in the classification results. We have studied the performance of several well-known kernels, including: Log-

TABLE III

COMPARISON BETWEEN AFFINE-INVARIANT AND LOG-EUCLIDEAN GAUSSIAN KERNELS USED IN THE PROPOSED APPROACH

| kernels | Depression/Normal | | ADHD/BMD | |
|---|---|---|---|---|
| | Eyes-open | Eyes-closed | Eyes-open | Eyes-closed |
| AIGK | 86.00 | 90.00 | 83.72 | 88.37 |
| LEGK | 76.00 | 80.00 | 74.42 | 76.74 |

Euclidean Gaussian kernel (LEGK), affine-invariant Gaussian kernel (AIGK), and Isometric kernel at concept-level (Table III). These comparisons are performed on the ADHD/BMD and depression/normal datasets in two eyes-open and eyes-closed resting conditions. The EEG signals are filtered in 3-50 Hz.

The concept-level positive definiteness of Log-Euclidean Gaussian kernel, leads to positive definiteness of its sum over different pairs of concepts in MI framework. Therefore, it provides a *pd* representation. Gaussian kernel based on the affine-invariant Riemannian metric is not *pd* in general.

Although the positive definiteness of similarity matrix, guarantees the convergence of kernel-based method to an optimal solution, but it does not necessarily lead to superior classification performance in comparison with a non-*psd* kernel. The superior results achieved by affine-invariant Gaussian kernel in this experiment confirm this fact. One interesting observation in this experiment is the positive definiteness of empirical affine-invariant Gaussian kernel achieved on ADHD/BMD and depression/normal datasets. This observation is confirmed by a recent work done by Feragen and Hauberg [48] which states that the affine-invariant Gaussian kernel has a high probability of being *pd* for a given data set in a large range of values.

We repeated the same experiment using Isometric kernel for SVM classifier which led to the overfitting to the training samples. This kernel resulted in about 50% negative eigen-fraction for indefinite similarity matrices for these datasets in eye-closed resting condition (-0.4950 for depression/normal and -0.4958 for ADHD/BMD datasets. The resulting poor generalization can be caused by ignorance of the 50% information that was conveyed by negative eigen components of the similarity matrix [34, 35]. It implies the necessity of considering the information of negative components and rectifying the matrix. As a future work we can consider the topological information conveyed by the negative components of the resulting representation.

Robustness of the proposed approach against noise was examined by adding the white Gaussian noise with different signal-to-noise ratios (SNR) to signal (10db, 5db, 0db). We tried this experiment on both ADHD/BMD and depression/normal datasets in eyes-closed resting conditions. The signals involved in this experiment are filtered in 3-50Hz and leave-one-out cross-validation is used for evaluation. The performance of the proposed method on signals which are contaminated by white Gaussian noise with different SNR levels are reported in Table IV. As the experimental evidences confirm, the proposed approach is relatively robust in noisy conditions. The performance of the method decreases about 2 to 4 percent where the noise is considerable (0db) and about 2

TABLE IV

PERFORMANCE OF MI-ADAPTIVE-COV+SVM ON ADHD/BMD AND DEPRESSION/NORMAL DATASETS IN EYES-OPEN AND EYES-CLOSED RESTING CONDITIONS, THE DATASETS ARE CONTAMINATED BY WHITE GAUSSIAN NOISE WITH DIFFERENT SNR LEVELS.

| | 10db | 5db | 0db |
|---|---|---|---|
| Depression/normal (eyes-closed) | 88.00 | 88.00 | 86.00 |
| Depression/normal (eyes-open) | 84.00 | 82.00 | 82.00 |
| ADHD/BMD (eyes-closed) | 86.05 | 86.05 | 86.05 |
| ADHD/BMD (eye-open) | 81.40 | 81.40 | 79.07 |

percent where the SNR is equal to 10db.

## V. CONCLUSION

In this paper, we proposed a new representation for describing EEG signals. In this representation we apply the MI framework to consider the non-stationarity of the EEG signal. The concepts in this representation are described by the empirical covariance matrix of EEG segments. These segments are the result of the adaptive segmentation of the signal and all the analysis are performed in Riemannian framework. Using MI kernel for describing EEG signal is a suitable choice which magnifies the discriminative information by automatically attenuating the noise contribution (i.e. considering it as a constant offset in similarity measure). Experimental evidences confirm the significant superiority of this representation in EEG-based CAD applications. This superiority comes from considering the non-stationarity of the EEG signal without the need for explicit induction of EEG segments, using covariance matrix for describing segments, determining the boundaries of the segments adaptively and the benefits of analysis of covariance matrices in Riemannian framework. The importance of selection of an appropriate concept-level kernel on the performance of the diagnosis system and the importance of rectifying the non-*pd* representation matrix, where the negative eigen-fraction of representation matrix is considerable, motivate us to investigate these two problems as future works.

ACKNOWLEDGMENT

The first author specially acknowledges Dr. Mathieu Salzmann, Dr. Saeed Sanei, and Dr. Reza Boostani for their helps and useful discussions.

REFERENCES

[1]  A. P. Association, *Diagnostic and statistical manual of mental disorders, fourth edition: DSM-IV-TR®*. American Psychiatric Association, 2000. [Online]. Available: https://books.google.com/books/about/Diagnostic_and_Statistical_Manual_of_Men.html?id=3SQrtpnHb9MC. Accessed: Nov. 12, 2016.

[2]   U. R. Acharya, V. K. Sudarshan, H. Adeli, J. Santhosh, J. E. W. Koh, and A. Adeli, "Computer-aided diagnosis of depression using EEG signals," *European Neurology*, vol. 73, no. 5-6, pp. 329–336, May 2015.

[3] D. V. Iosifescu *et al.*, "Frontal EEG predictors of treatment outcome in major depressive disorder," *European Neuropsychopharmacology*, vol. 19, no. 11, pp. 772–777, Nov. 2009.

[4] C. Salustri *et al.*, "Cortical excitability and rest activity properties in patients with depression," *Journal of psychiatry & neuroscience: JPN.*, vol. 32, no. 4, pp. 259–66, Jul. 2007. [Online]. Available: https://www.ncbi.nlm.nih.gov/pubmed/17653294. Accessed: Nov. 12, 2016.

[5] A. A. Fingelkurts, H. Rytsälä, K. Suominen, E. Isometsä, and S. Kähkönen, "Composition of brain oscillations in ongoing EEG during major depression disorder," *Neuroscience Research*, vol. 56, no. 2, pp. 133–144, Oct. 2006.

[6] H. Adeli, Z. Zhou, and N. Dadmehr, "Analysis of EEG records in an epileptic patient using wavelet transform," *Journal of Neuroscience Methods*, vol. 123, no. 1, pp. 69–87, Feb. 2003.

[7] U. R. ACHARYA, C. K. CHUA, T.-C. LIM, DORITHY, and J. S. SURI, "AUTOMATIC IDENTIFICATION OF EPILEPTIC EEG SIGNALS USING NONLINEAR PARAMETERS," *Journal of Mechanics in Medicine and Biology*, vol. 09, no. 04, pp. 539–553, Dec. 2009.

[8] C. Kuang Chua, "Cardiac health diagnosis using higher order spectra and support vector machine," *The Open Medical Informatics Journal*, vol. 3, no. 1, pp. 1–8, 2010.

[9] R. J. MARTIS *et al.*, "APPLICATION OF EMPIRICAL MODE DECOMPOSITION (EMD) FOR AUTOMATED DETECTION OF EPILEPSY USING EEG SIGNALS," *International Journal of Neural Systems*, vol. 22, no. 06, p. 1250027, Dec. 2012.

[10] Q. YUAN, W. ZHOU, S. YUAN, X. LI, J. WANG, and G. JIA, "EPILEPTIC EEG CLASSIFICATION BASED ON KERNEL SPARSE REPRESENTATION," *International Journal of Neural Systems*, vol. 24, no. 04, p. 1450015, Jun. 2014.

[11] H. Adeli, S. Ghosh-Dastidar, and N. Dadmehr, "A spatio-temporal wavelet-chaos methodology for EEG-based diagnosis of Alzheimer's disease," *Neuroscience Letters*, vol. 444, no. 2, pp. 190–194, Oct. 2008.

[12] R. Boostani, K. Sadatnezhad, and M. Sabeti, "An efficient classifier to diagnose of schizophrenia based on the EEG signals," *Expert Systems with Applications*, vol. 36, no. 3, pp. 6492–6499, Apr. 2009.

[13] S. D. PUTHANKATTIL and P. K. JOSEPH, "CLASSIFICATION OF EEG SIGNALS IN NORMAL AND DEPRESSION CONDITIONS BY ANN USING RWE AND SIGNAL ENTROPY," *Journal of Mechanics in Medicine and Biology*, vol. 12, no. 04, p. 1240019, Sep. 2012.

[14] O. FAUST, P. C. A. ANG, S. D. PUTHANKATTIL, and P. K. JOSEPH, "DEPRESSION DIAGNOSIS SUPPORT SYSTEM BASED ON EEG SIGNAL ENTROPIES," *Journal of Mechanics in Medicine and Biology*, vol. 14, no. 03, p. 1450035, Jun. 2014.

[15] B. Hosseinifard, M. H. Moradi, and R. Rostami, "Classifying depression patients and normal subjects using machine learning techniques and nonlinear features from EEG signal," *Computer Methods and Programs in Biomedicine*, vol. 109, no. 3, pp. 339–345, Mar. 2013.

[16] M. Congedo, A. Barachant, and A. Andreev, "Title: A new generation of brain-computer interface based on Riemannian geometry," 2013. [Online]. Available: https://arxiv.org/abs/1310.8115. Accessed: Nov. 12, 2016.

[17] X. Pennec, P. Fillard, and N. Ayache, "A Riemannian framework for Tensor computing," *International Journal of Computer Vision*, vol. 66, no. 1, pp. 41–66, Jan. 2006.

[18] A. Y. Kaplan, A. A. Fingelkurts, S. V. Borisov, and B. S. Darkhovsky, "Nonstationary nature of the brain activity as revealed by EEG/MEG: Methodological, practical and conceptual challenges," *Signal Processing*, vol. 85, no. 11, pp. 2190–2212, Nov. 2005.

[19] S. Sanei and J. A. Chambers, *EEG signal processing*. Chichester, England: Wiley-Blackwell (an imprint of John Wiley & Sons Ltd), 2007.

[20] T. G. Dietterich, R. H. Lathrop, and T. Lozano-Pérez, "Solving the multiple instance problem with axis-parallel rectangles," *Artificial Intelligence*, vol. 89, no. 1-2, pp. 31–71, Jan. 1997.

[21] J. M. Lee, *Riemannian Manifolds: An introduction to curvature*. Springer Science & Business Media, 2006. [Online]. Available: https://books.google.com/books?hl=en&lr=&id=92PgBwAAQBAJ&oi=fnd&pg=PA1&dq=Riemannian+manifolds:+an+introduction+to+curvature+&ots=dBxU0H5hJG&sig=rwCpqZ_e7RlOas2LNIgSQRguhE4#v=onepage&q=Riemannian%20manifolds%3A%20an%20introduction%20to%20curvature&f=false. Accessed: Nov. 12, 2016.

[22] J. Jost, *Riemannian geometry and geometric analysis*. Springer Science & Business Media, 2013. [Online]. Available: https://books.google.com/books?id=HG3mCAAAQBAJ&printsec=frontcover&dq=Riemannian+geometry+and+geometric+analysis&hl=en&sa=X&ved=0ahUKEwjCxOvI-KPQAhUDWywKHenDCEMQ6AEIHDAA#v=onepage&q=Riemannian%20geometry%20and%20geometric%20analysis&f=false. Accessed: Nov. 12, 2016.

[23] T. Lin and H. Zha, "Riemannian manifold learning," *IEEE Transactions on Pattern Analysis and Machine Intelligence*, vol. 30, no. 5, pp. 796–809, May 2008.

[24] V. Arsigny, P. Fillard, X. Pennec, and N. Ayache, "Log-Euclidean metrics for fast and simple calculus on diffusion tensors," *Magnetic Resonance in Medicine*, vol. 56, no. 2, pp. 411–421, 2006.

[25] I. L. Dryden, A. Koloydenko, and D. Zhou, "Non-Euclidean statistics for covariance matrices, with applications to diffusion tensor imaging," *The Annals of Applied Statistics*, vol. 3, no. 3, pp. 1102–1123, Sep. 2009.

[26] S. Andrews, I. Tsochantaridis, and T. Hofmann, "Support vector machines for multiple-instance learning,". [Online]. Available: http://papers.nips.cc/paper/2232-support-vector-machines-for-multiple-instance-learning.pdf. Accessed: Nov. 12, 2016.

[27] J. Ramon, "Multi Instance Neural Networks," in *In Proceedings of the International Conference on Machine Learning 2000 Workshop on Attribute-Value and Relational Learning*, 2000. [Online]. Available: http://citeseerx.ist.psu.edu/viewdoc/summary?doi=10.1.1.43.682. Accessed: Nov. 12, 2016.

[28] H. Blockeel, D. Page, and A. Srinivasan, "Multi-instance tree learning," ACM, 2005, pp. 57–64. [Online]. Available: http://dl.acm.org/citation.cfm?id=1102359. Accessed: Nov. 12, 2016.

[29] P. Auer and R. Ortner, "A boosting approach to multiple instance learning," in *Lecture Notes in Computer Science*, Springer Science + Business Media, 2004, pp. 63–74.

[30] A. Barachant, A. Andreev, and M. Congedo, "The Riemannian potato: An automatic and adaptive artifact detection method for online experiments using Riemannian geometry," 2013, pp. 19–20. [Online]. Available: https://hal.archives-ouvertes.fr/hal-00781701/. Accessed: Nov. 12, 2016.

[31] S. Jayasumana, R. Hartley, M. Salzmann, H. Li, and M. Harandi, "Kernel methods on Riemannian Manifolds with Gaussian RBF kernels," *IEEE Transactions on Pattern Analysis and Machine Intelligence*, vol. 37, no. 12, pp. 2464–2477, Dec. 2015.

[32] I. Jolliffe, T. F. Cox, and M. A. A. Cox, "Multidimensional scaling," *Technometrics*, vol. 38, no. 4, p. 403, Nov. 1996.

[33] V. De Silva and J. B. Tenenbaum, "Global versus local methods in Nonlinear Dimensionality reduction,". [Online]. Available: http://machinelearning.wustl.edu/mlpapers/paper_files/AA28.pdf. Accessed: Nov. 12, 2016.

[34] E. Pekalska, P. Paclik, and R. P. W. Duin, "A generalized kernel approach to Dissimilarity-based classification," *Journal of Machine Learning Research*, vol. 2, no. Dec, pp. 175–211, 2001. [Online]. Available: http://www.jmlr.org/papers/v2/pekalska01ar1.html. Accessed: Nov. 12, 2016.

[35] E. Pękalska and R. P. W. Duin, "Dissimilarity representations allow for building good classifiers," *Pattern Recognition Letters*, vol. 23, no. 8, pp. 943–956, Jun. 2002.

[36] K. Sadatnezhad, R. Boostani, and A. Ghanizadeh, "Proposing an adaptive mutation to improve XCSF performance to classify ADHD and BMD patients," *Journal of Neural Engineering*, vol. 7, no. 6, p. 066006, Nov. 2010.

[37] R. Kazemi *et al.*, "Electrophysiological correlates of bilateral and unilateral repetitive transcranial magnetic stimulation in patients with bipolar depression," *Psychiatry Research*, vol. 240, pp. 364–375, Jun. 2016.

[38] WHO, "Official WHO health days," in *World Health Organization*, World Health Organization, 2016. [Online]. Available: http://www.who.int/mediacentre/events/annual/world_suicide_prevention_day/en. Accessed: Nov. 12, 2016.

[39] T. Higuchi, "Approach to an irregular time series on the basis of the fractal theory," *Physica D: Nonlinear Phenomena*, vol. 31, no. 2, pp. 277–283, Jun. 1988.

[40] P. Stoica and R. L. Moses, "Introduction to Spectral Analysis," 1997. [Online]. Available: http://tocs.ulb.tu-darmstadt.de/60534176.pdf. Accessed: Nov. 12, 2016.

[41] M. Sabeti, S. Katebi, and R. Boostani, "Entropy and complexity measures for EEG signal classification of schizophrenic and control participants," *Artificial Intelligence in Medicine*, vol. 47, no. 3, pp. 263–274, Nov. 2009.

[42] S. Expert, "Spectral analysis of EEG signals during hypnosis - Asociația Română de Hipnoză," Aug. 2013. [Online]. Available: http://asociatiaromanadehipnoza.ro/spectral-analysis-of-eeg-signals-during-hypnosis-2/. Accessed: Nov. 12, 2016.

[43] H. Adeli, S. Ghosh-Dastidar, and N. Dadmehr, "A Wavelet-Chaos methodology for analysis of EEGs and EEG Subbands to detect seizure and epilepsy," *IEEE Transactions on Biomedical Engineering*, vol. 54, no. 2, pp. 205–211, Feb. 2007.

[44] J. A. K. Suykens and J. Vandewalle, *Neural Processing Letters*, vol. 9, no. 3, pp. 293–300, 1999.

[45] C.-C. Chang and C.-J. Lin, "LIBSVM," *ACM Transactions on Intelligent Systems and Technology*, vol. 2, no. 3, pp. 1–27, Apr. 2011.

[46] R. Bhatia and J. Holbrook, "Riemannian geometry and matrix geometric means," *Linear Algebra and its Applications*, vol. 413, no. 2-3, pp. 594–618, Mar. 2006.

[47] J. Foulds and E. Frank, "A review of multi-instance learning assumptions," *The Knowledge Engineering Review*, vol. 25, no. 01, p. 1, Mar. 2010.

[48] A. Feragen, A. Ku, and S. Dk, "Open problem: Kernel methods on manifolds and metric spaces what is the probability of a positive definite geodesic exponential kernel? Søren Hauberg," vol. 49, pp. 1–4, 2016. [Online]. Available: http://www.jmlr.org/proceedings/papers/v49/feragen16.pdf. Accessed: Nov. 12, 2016.

[49] G. Wu, E. Y. K. Chang, and Z. Zhang, "An analysis of transformation on non-positive semidefinite similarity matrix for kernel machines," in *In Proceedings of the 22nd International Conference on Machine Learning*. [Online]. Available: http://citeseerx.ist.psu.edu/viewdoc/summary?doi=10.1.1.133.4077. Accessed: Nov. 12, 2016.

[50] J. A. Lee and M. Verleysen, *Nonlinear dimensionality reduction*. New York, NY: Springer-Verlag New York, 2007.

[51] T. Zhang and W. Chen, "LMD based features for the automatic seizure detection of EEG signals using SVM," *IEEE Transactions on Neural Systems and Rehabilitation Engineering*, pp. 1–1, 2016.

[52] D. Garrett, D. A. Peterson, C. W. Anderson, and M. H. Thaut, "Comparison of linear, nonlinear, and feature selection methods for eeg signal classification," *IEEE Transactions on Neural Systems and Rehabilitation Engineering*, vol. 11, no. 2, pp. 141–144, Jun. 2003.